\documentclass[lettersize,journal]{IEEEtran}

\usepackage{amsmath,amsfonts}
\usepackage{algorithmic}
\usepackage{array}
\usepackage{caption}
\usepackage{subcaption}
\usepackage{textcomp}
\usepackage{stfloats}
\usepackage{url}
\usepackage{verbatim}
\usepackage{graphicx}
\usepackage[switch]{lineno}
\def\BibTeX{{\rm B\kern-.05em{\sc i\kern-.025em b}\kern-.08em
    T\kern-.1667em\lower.7ex\hbox{E}\kern-.125emX}}
\usepackage{balance}

\usepackage{booktabs}
\usepackage{multirow}
\usepackage{xcolor}
\usepackage[pagebackref,breaklinks,colorlinks,citecolor=blue]{hyperref}
\usepackage[capitalise]{cleveref}
\usepackage{orcidlink}
\usepackage{hyperref}
\begin{document}
\title{\textcolor{blue}{VS3R}: Robust Full-frame \textcolor{blue}{V}ideo \textcolor{blue}{S}tabilization \\via Deep \textcolor{blue}{3}D \textcolor{blue}{R}econstruction} 

\author{Muhua Zhu~\orcidlink{0009-0005-9049-2321}, 
Xinhao Jin~\orcidlink{0009-0006-5722-2896}, 
Xinping Wang~\orcidlink{0009-0008-3022-5768}, 
Yu Zhang~\orcidlink{0009-0005-2353-988X}, 
Yifei Xue~\orcidlink{0000-0002-4443-4367}, 
Tie Ji~\orcidlink{0009-0005-0822-0600}, 
and Yizhen Lao~\orcidlink{0000-0002-6284-1724}
\thanks{Muhua Zhu and Xinhao Jin contributed equally to this work. (Corresponding authors: Yizhen Lao.)}
\thanks{The authors are with Hunan University, Changsha 410082, China (e-mail: casmyzhu@hnu.edu.cn; jinxinhao@hnu.edu.cn; wxpsb@hnu.edu.cn; yuzhang@hnu.edu.cn; iflyhsueh@gmail.com; jietie\_hnu@163.com; yizhenlao@hnu.edu.cn).}
\thanks{Manuscript received XXXX, 2026; revised XXXX, 2026.}
}

\markboth{IEEE Transactions on Multimedia,~Vol.~XX, No.~XX, XXXX~2026}%
{Zhu \MakeLowercase{\textit{et al.}}: VS3R: Robust Full-frame Video Stabilization via Deep 3D Reconstruction}
\maketitle

\begin{figure*}
  \centering
  \includegraphics[width=\textwidth]{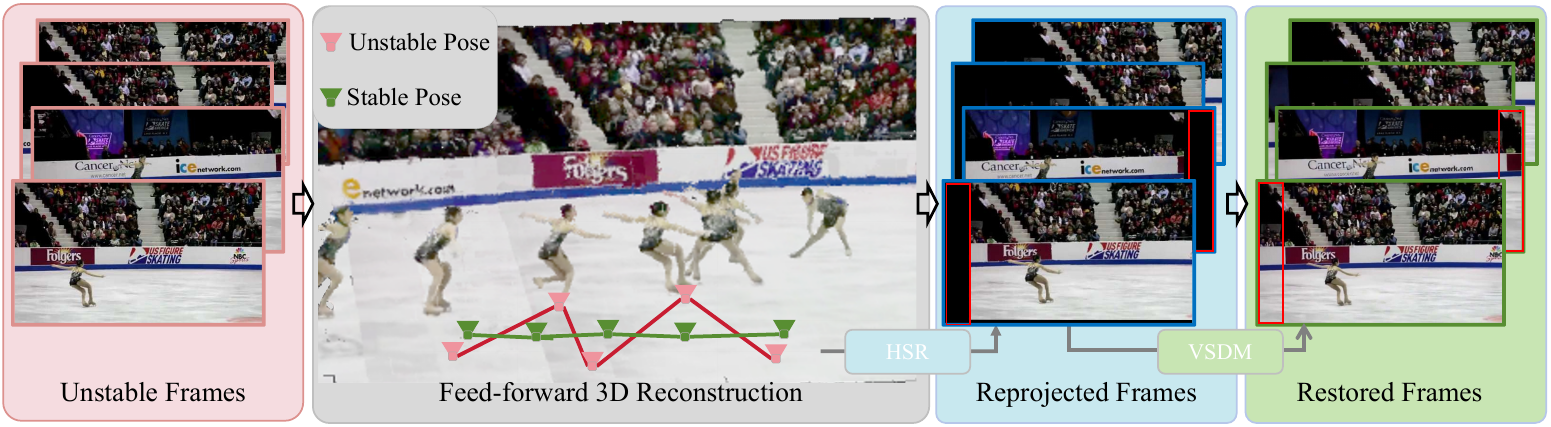}
  \caption{We study the task of video stabilization from a novel 3D perspective. Given an uncalibrated video, we first reconstruct the local 3D scene using a feed-forward model. We then employ a hybrid rendering strategy, fusing semantic and geometric cues to synthesize stabilized frames. Finally, a dual-stream video diffusion model performs full-frame completion and refinement, producing high-fidelity, temporally coherent video.}
  \label{fig:teaser}
\end{figure*}

\begin{abstract}
Video stabilization aims to mitigate camera shake but faces a fundamental trade-off between geometric robustness and full-frame consistency. While 2D methods suffer from aggressive cropping, 3D techniques are often undermined by fragile optimization pipelines that fail under extreme motions. Novel view synthesis models suffer from structural artifacts and scale blindness. To bridge this gap, we propose \textcolor{blue}{VS3R}, a framework that synergizes feed-forward 3D reconstruction with generative video diffusion. Our pipeline jointly estimates camera parameters, depth, and masks to ensure all-scenario reliability, and introduces a \textbf{Hybrid Stabilized Rendering (HSR)} module that fuses semantic and geometric cues to preliminarily address parallax occlusions caused by pose transformations while maintaining dynamic-static consistency. Finally, a \textbf{Video Stabilization-Driven Diffusion Model (VSDM)} leverages contextual information to restore disoccluded regions, jointly optimizing texture and temporal consistency. Collectively, VS3R achieves high-fidelity, full-frame stabilization across diverse camera models and significantly outperforms state-of-the-art methods in robustness and visual quality.
\end{abstract}

\begin{IEEEkeywords}
Video Stabilization, 3D Reconstruction, Diffusion Model, Deep Learning. 
\end{IEEEkeywords}

\section{Introduction}
Video stabilization aims to eliminate unintended camera shake induced by handheld shooting or vehicle-mounted platforms\cite{guilluy2021video}. A typical video stabilization pipeline comprises three core stages, namely, motion estimation of the unstable video, motion smoothing, and stabilized frame synthesis. Based on the underlying motion models employed for estimation, existing methods can be broadly categorized into 2D and 3D approaches.

\begin{figure*}
    \centering
    \includegraphics[width=0.8\textwidth]{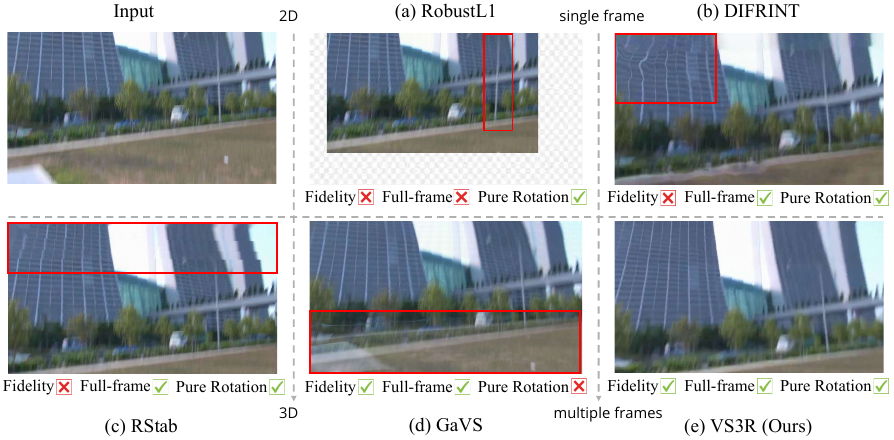}
    \caption{
\textbf{Comparison of video stabilization paradigms in rapid pure-rotation scene, characterized by extreme motions, blur, and severe cropping.} (a) 2D methods: Significant information loss due to aggressive cropping. (b) Learning-based: Geometric distortions and temporal flickering. (c) Completion-based: Failure in maintaining structural integrity. (d) SfM-based: Pose estimation failure under geometric degeneracy. (e) Ours: Full-frame, high-fidelity stabilization with robust geometric and temporal consistency.
    }
    \label{fig:related_work_fig}
\end{figure*}

Traditional \textbf{2D-based methods} typically employ planar transformations, such as affine~\cite{grundmann2011auto}, homography~\cite{matsushita2006full,liu2013bundled,liu2017codingflow}, or mesh-based warping~\cite{liu2013bundled, liu2016meshflow, wang2018deep}, for frame alignment. While recent learning-based approaches attempt to implicitly estimate motion-compensated warping fields~\cite{Yu_2019_CVPR, Yu_2020_CVPR, Zhang_2023_ICCV, Liu_2025_TPAMI, DIFRINT}, they often struggle to generalize across diverse and complex scenarios due to data limitations. More critically, purely 2D stabilization methods lack physical constraints derived from the underlying 3D scene geometry. Consequently, these methods frequently fail in environments characterized by parallax, leading to severe structural distortions and temporal inconsistency as shown in Fig.~\ref{fig:related_work_fig}\textcolor{red}{b}. To hide these artifacts, 2D methods inevitably resort to aggressive cropping, resulting in a significant loss of field-of-view~\cite{grundmann2011auto, liu2017codingflow} as shown in Fig.~\ref{fig:related_work_fig}\textcolor{red}{a}.

To preserve structural integrity, Some \textbf{2.5D-based methods} incorporate depth cues~\cite{lee20213d} or subspace constraints~\cite{liu2011subspace} to enforce geometric consistency. Despite their ability to preserve scene structure, these 3D-aware methods frequently fail to synthesize full-frame content, often leaving behind projection artifacts or incomplete boundaries. Recently, \textbf{3D-based approaches} have emerged, leveraging paradigms such as NeRF~\cite{peng20243d} or 3D Gaussian Splatting~\cite{you2025gavs} within reconstruct-and-render pipelines. However, these 3D methods heavily rely on standard Structure-from-Motion (SfM) for camera pose and depth estimation~\cite{peng20243d, you2025gavs}. SfM-based frameworks are inherently fragile in ill-posed scenarios, such as pure rotation or motion blur, where tracking often fails or suffers from severe scale drift due to geometric degeneracy, as depicted in Fig.~\ref{fig:related_work_fig}\textcolor{red}{d}~\cite{you2025gavs}. Furthermore, existing 3D techniques frequently struggle to synthesize full-frame content, often leaving behind projection artifacts or incomplete boundaries due to their limited handling of dynamic objects.

Some novel view synthesis works claim to be applicable to downstream video stabilization tasks, yet they fail to align with the specific objectives of video stabilization. ~\cite{yu2025trajectorycrafter} assumes a static input camera by assigning fixed poses to all frames. Additionally, resolving its unnormalized translation scales from monocular depth requires heavy offline preprocessing with target poses and point clouds. Furthermore, monocular depth estimation often introduces flying pixels in regions of depth discontinuity, thereby preventing it from properly handling occlusion relationships. ~\cite{ren2025gen3c} shares this monocular depth scale ambiguity, making it similarly inapplicable. 
~\cite{jeong2025reangle} relies solely on a single image, specifically the first frame, for inference and cannot take multiple images as input. ~\cite{bai2025recammaster} implicitly uses camera poses for video rendering, but without explicit geometric modeling, it struggles to handle parallax. ~\cite{yang2026neoverse} similarly employs a paradigm combining multi-view reconstruction and generation. However, the inherent scale-blindness and severe aliasing of unoptimized 3DGS make it difficult to handle complex camera dynamics, such as rapid focal length variations (Fig.~\ref{fig:neo_our1}). Furthermore, over-relying on the generative model to resolve all degradations often leads to hallucinations. Under poor rendering conditions or when aggressive stabilization creates large missing boundaries, the model generates content entirely inconsistent with the original scene (\cref{fig:neo_our2,fig:neo_our3}).

\textbf{\textit{In summary}}, current stabilization paradigms face a fundamental trade-off between geometric robustness and full-frame consistency. While 2D-based temporal aggregation~\cite{DIFRINT, Zhao_2023_ICCV, Liu_2021_ICCV, peng20243d} attempts to fill disoccluded regions to mitigate cropping-induced information loss, it frequently fails to maintain structural integrity, particularly under severe cropping (Fig.~\ref{fig:related_work_fig}\textcolor{red}{c}). Conversely, although 3D-aware strategies incorporate depth priors~\cite{lee20213d, liu2012video} or auxiliary sensors~\cite{li2022deep, shi2022deep} to enforce geometric fidelity, they often struggle with full-frame synthesis and projection artifacts. Most critically, emerging NeRF~\cite{peng20243d} and 3DGS~\cite{you2025gavs} paradigms typically rely on decoupled pipelines, combining SfM with monocular depth, which triggers a cascading amplification of errors. This modularity renders them fragile under geometric degeneracies (e.g., pure rotation) or aggressive motion scenarios, where traditional SfM frameworks suffer from tracking failures and learning-based estimators often drift. Furthermore, novel view synthesis paradigms heavily rely on generative models while neglecting rendering optimization, making them prone to hallucinating scene-irrelevant content when applied to video stabilization.

\textbf{Thus,} there remains a critical need for a unified stabilization paradigm that concurrently delivers all-scenario robustness, high-fidelity full-frame synthesis, and temporal consistency across diverse scenes and complex camera motions.

To address the aforementioned challenges, we propose \textbf{\textcolor{blue}{VS3R}} (Fig.~\ref{fig:pipeline}), a novel and robust framework for video stabilization that leverages deep 3D reconstruction in conjunction with video diffusion models. \textbf{\textit{(1)}} We first apply the feed-forward scene reconstruction model~\cite{hu2025vggt4d} to uncalibrated video frames, jointly recovering camera intrinsics, extrinsics, dynamic masks, and depth maps. To prevent global prediction drift and memory explosion typically associated with long-range sequences, we process the video sequence using a sliding window approach. Subsequently, we employ a Gaussian filter to smooth the camera trajectory. \textbf{\textit{(2)}} Utilizing these optimized poses, we perform hybrid rendering of both static and dynamic points within the window to ensure geometric consistency and to initially address parallax occlusion. \textbf{\textit{(3)}} To achieve full-frame restoration, we introduce a Video Stabilization-Driven Diffusion Model (VSDM). By decoupling scene aggregation and structure-texture optimization via a Mixture-of-Experts (MoE) architecture, it effectively leverages adjacent frames to fill missing details, resolve parallax occlusions, and rectify artifacts in the rendered sequence. 
Experimental results demonstrate that \textcolor{blue}{VS3R} outperforms existing methods both qualitatively and quantitatively, while simultaneously achieving full-frame synthesis, content and geometric preservation, and robustness across diverse scenes. \textcolor{blue}{To facilitate reproducibility, the source code of VS3R is included in the supplementary material and is publicly available.}
In summary, our contributions are summarized as:

\begin{itemize}
  \item We propose a deep 3D reconstruction-based stabilization pipeline capable of generating full-frame videos with content, geometric, and temporal consistency under diverse and challenging camera motions. 
  \item We introduce a Hybrid Stabilized Rendering (HSR) module that synergizes semantic and geometric cues to ensure consistency, complemented by a Video Stabilization-Driven Diffusion Model (VSDM) that restores disoccluded regions and rectifies artifacts via temporal aggregation, collectively achieving high-fidelity, full-frame stabilization without the need for aggressive cropping.
  \item Extensive experiments on public benchmarks demonstrate that VS3R significantly outperforms SOTA 2D and 3D methods. Our framework's effectiveness is validated through quantitative metrics, qualitative comparisons, and a blind user study, showing superior robustness under extreme motions.
\end{itemize}
\section{Related Work}
\noindent $\bullet$\textbf{2D-based Video Stabilization.} Earlier 2D methods typically model camera motion as a series of 2D transformations on the image plane, such as affine \cite{grundmann2011auto}, homography \cite{matsushita2006full,goldstein2012video,liu2013bundled,liu2017codingflow}, feature tracks~\cite{yu2018selfie,koh2015video,liu2014steadyflow,lee2009video}, mesh flow \cite{liu2013bundled, liu2016meshflow,liu2023content} and dense flow~\cite{chen2021pixstabnet,DIFRINT,liu2021hybrid,zhao2020pwstablenet}. While efficient, these 2D-based approaches lack explicit 3D scene modeling, making them prone to geometric distortions in the presence of complex parallax. Furthermore, to avoid disoccluded regions, these methods inevitably rely on aggressive cropping, resulting in significant loss of field-of-view (FoV). Recent learning-based 2D methods \cite{Yu_2020_CVPR, Zhang_2023_ICCV, Liu_2025_TPAMI} attempt to implicitly learn warping fields, yet they still struggle with the fundamental trade-off between stability and content preservation.

\noindent  $\bullet$\textbf{2.5D and 3D-based Video Stabilization.} To preserve structural integrity, some 2.5D-based approaches utilize additional hardware such as depth cameras \cite{liu2012video} or IMUs \cite{li2022deep, shi2022deep} to assist in recovering geometry. Some methods leverage depth cues~\cite{lee20213d} or subspace constraints~\cite{liu2011subspace} to enforce 3D-like consistency without full reconstruction. Recently, 3D-based methods have leveraged NeRF~\cite{peng20243d} or 3D Gaussian Splatting~\cite{you2025gavs} to maintain geometric consistency through a reconstruct-and-render pipeline. However, their heavy reliance on standard SfM limits their robustness and generalization in ill-posed scenarios.

\noindent  $\bullet$\textbf{Novel view synthesis} Recent advances in diffusion models~\cite{ho2020denoising,rombach2022high} have significantly propelled 3D/4D reconstruction and conditional video generation. Several methods~\cite{yu2025trajectorycrafter, ren2025gen3c, jeong2025reangle} adopt a reconstruction-generation paradigm, leveraging monocular depth for pose-controlled novel view rendering, followed by generative models for visual completion. However, monocular depth estimation often introduces flying pixels at depth boundaries, failing to properly handle occlusions. Alternatively, while ~\cite{yang2026neoverse} effectively addresses parallax via multi-view reconstruction, its heavy reliance on the generative model often produces scene-inconsistent content under poor rendering conditions or within the missing boundary regions induced by aggressive stabilization.
\section{Methodology}
\label{sec:methodology}
\subsection{pipeline}
\begin{figure*}[ht]
    \centering
    \includegraphics[width=\textwidth]{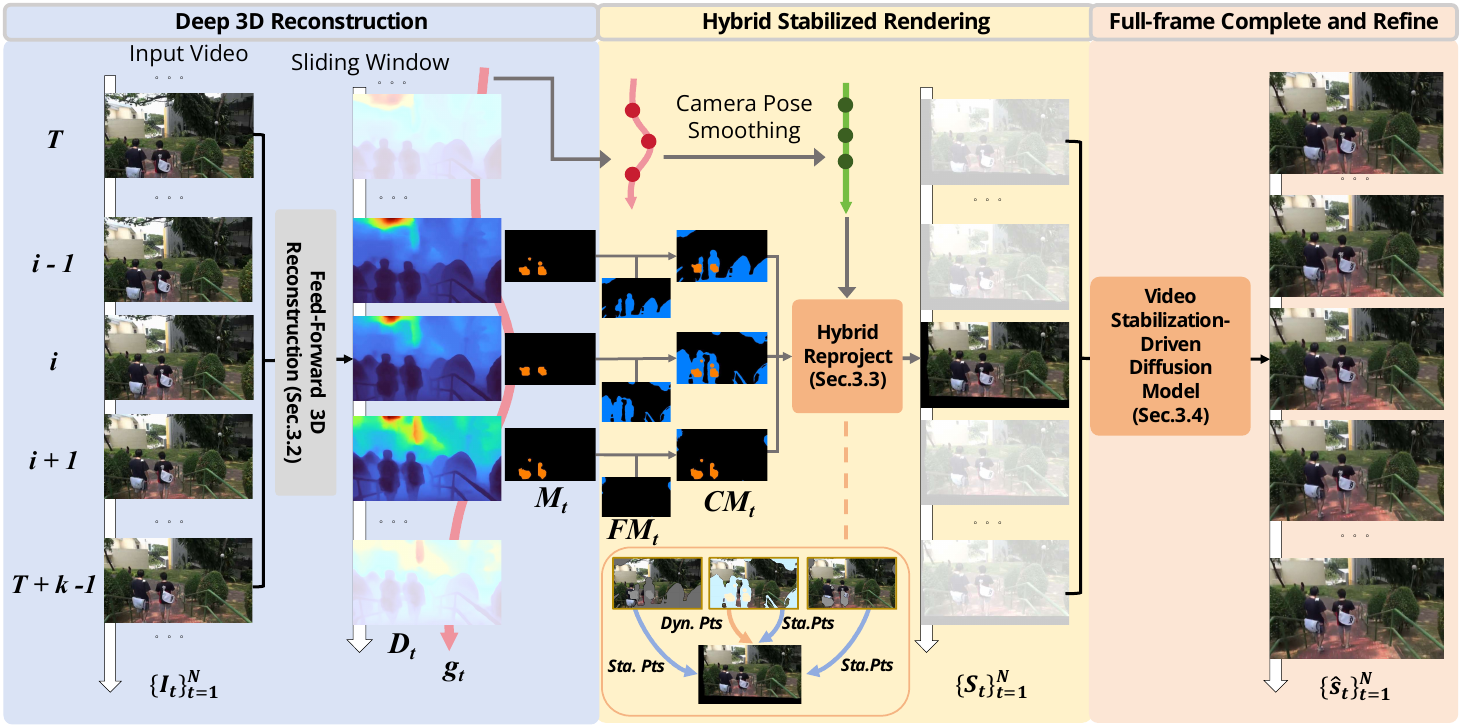}
    \caption{
    \textbf{Overview of the \textcolor{blue}{VS3R} pipeline.} \textcolor{blue}{VS3R} follows a "reconstruct-smooth-refine" paradigm: 
    (1) Deep 3D Reconstruction (Sec.~\ref{subsec:scene reconstruction}): Estimates camera parameters $g_t$, semantic masks $M_t$, and depth $D_t$ from uncalibrated video via a feed-forward model~\cite{hu2025vggt4d}. 
    (2) Hybrid Stabilized Rendering (Sec.~\ref{subsec:HSR}): Refines dynamic mask $CM_t$ by merging $M_t$ with geometric mask $FM_t$, then renders stabilized frames $S_t$ along the smoothed trajectory. 
    (3) Full-frame Refinement (Sec.~\ref{subsec:DVDM}): A Video Stabilization-Driven Diffusion Model(VSDM) restores disoccluded regions and rectifies artifacts to produce high-fidelity, temporally coherent frames $\hat{S}_t$.
    }  
    \label{fig:pipeline}
\end{figure*}
Fig.~\ref{fig:pipeline} illustrates the overall architecture of our \textcolor{blue}{VS3R} framework. To overcome the limitations of traditional SfM-based methods and 2D warping, \textcolor{blue}{VS3R} adopts a robust "reconstruct-smooth-refine" paradigm consisting of the following three core steps: \textbf{(1) Deep 3D Scene Reconstruction}. Given an uncalibrated unstable video sequence, we first employ a feed-forward deep reconstruction model~\cite{hu2025vggt4d} to jointly recover camera intrinsics, extrinsics, dynamic masks, and depth maps, as detailed in Sec.~\ref{subsec:scene reconstruction}. \textbf{(2) Hybrid Stabilized Rendering (HSR)}. To produce a stabilized video, we apply a Gaussian filter to the estimated camera trajectory to obtain a smoothed path. Subsequently, we perform hybrid rendering of both static scene elements and dynamic objects within the window, as elaborated in Sec.~\ref{subsec:HSR}. \textbf{(3)Full-frame Completion and Refinement.} The preliminary rendered frames often suffer from cropping artifacts and disoccluded regions at the boundaries. To achieve full-frame stabilization, we introduce a Video Stabilization-Driven Diffusion Model (VSDM),as depicted in Sec.~\ref{subsec:DVDM}.

\subsection{Deep 3D Reconstruction}
\label{subsec:scene reconstruction}

Unlike traditional 3D stabilization methods that rely on fragile Structure-from-Motion (SfM) optimization (Fig.~\ref{fig:related_work_fig}), and existing 3DGS approaches that fail to effectively render certain motions such as drastic focal length changes as shown in ~\cref{fig:neo_our1}, we adopt a feed-forward deep reconstruction paradigm VGGT4D~\cite{hu2025vggt4d} to ensure robustness in challenging scenarios. As shown in Fig.~\ref{fig:pipeline}, to facilitate long-range processing while mitigating global drift, the input RGB sequence $\mathcal{I} = \{I_i\}_{i=1}^N$ is handled via a sliding window approach. Specifically, we leverage VGGT4D~\cite{hu2025vggt4d} to jointly estimate camera parameters, scene structure, and temporal correspondences. The reconstruction process for a given window is formulated as:

\begin{equation}
    \mathcal{F} \left( \{I_i\}_{i=T}^{T+W-1}\right) = \left\{ \mathbf{g}_i, D_i, P_i, M_i\right\}_{i=T}^{T+W-1},
    \label{eq:reconstruction}
\end{equation}

where $W$ denotes the window size. For each frame $i$, $\mathbf{g}_i$ represents the camera's intrinsic $K_i$ and extrinsic parameters $T_i$, $D_i$ is the predicted per-pixel depth map, $P_i$ signifies the corresponding 3D point map, and $M_i$ represents the \textbf{semantic-driven dynamic mask}. However, while the feed-forward model leverages deep semantic priors to estimate object-level motion, these masks often exhibit inaccuracies or temporal inconsistencies, potentially leading to severe visual artifacts during the hybrid rendering process. The subsequent section will detail the HSR module, which utilizes these predictions to eliminate the aforementioned artifacts and ensure geometric consistency.

\subsection{Hybrid Stabilized Rendering}
\label{subsec:HSR}
\noindent \textbf{Camera Path Smoothing.}
To derive stabilized camera trajectories from the raw parameters $\mathbf{g}_i$, following~\cite{lee20213d}, we apply temporal Gaussian smoothing to both translation and rotation components:
\begin{equation}
    \hat{\mathbf{g}}_k = \sum_{i=k-r}^{k+r} \omega(i,k) \mathbf{g}_i
\end{equation}
\begin{equation}
    \omega(i,k) = \exp \left( -\frac{(i-k)^2}{2\sigma^2} \right)
\end{equation}
where $r$ denotes the temporal radius and $\omega(i,k)$ represents the normalized Gaussian weights. For rotation, filtering is performed in unit quaternion space with sign-alignment to preserve motion continuity. By tuning the bandwidth parameter $\sigma$, our system synthesizes videos with varying degrees of stabilization.\\

\noindent \textbf{Hybrid Dynamic Mask.}
To suppress potential artifacts from moving objects, we refine the dynamic mask by merging the semantic-aware dynamic mask $M_t$ and geometric dynamic mask $FM_t$. To obtain $FM_t$, we first define the induced rigid flow $\mathbf{f}_{r}^{t}$ at pixel $\mathbf{p}$. Intuitively, $\mathbf{f}_{r}^{t}$ represents the expected pixel motion caused solely by the camera's ego-motion, assuming the underlying 3D scene is completely stationary. This projected motion field under the static scene assumption is formulated as::
\begin{equation}
    \mathbf{f}_{r}^{t}(\mathbf{p}) = K \left[ R_{t+1}^{-1} \left( (R_t K^{-1} \mathbf{p} D_t + \mathbf{t}_t) - \mathbf{t}_{t+1} \right) \right] - \mathbf{p}
\end{equation}
where $K$ denotes the camera intrinsic, and $\{R_t, \mathbf{t}_t\}$ represents the camera rotation and translation that transform a point from the camera coordinate system to the world coordinate system. $D_t$ is the depth of frame $t$. The geometric flow mask $FM_t$ is then determined by thresholding the $\ell_2$-norm of the residual between the observed optical flow $\mathbf{f}_{o}^{t}$, which is robustly estimated using the pre-trained RAFT Large model~\cite{teed2020raft}, and the induced rigid motion $\mathbf{f}_{r}^{t}$:
\begin{equation}
    FM_t(\mathbf{p}) = \mathbb{I} \left( \| \mathbf{f}_{o}^{t}(\mathbf{p}) - \mathbf{f}_{r}^{t}(\mathbf{p}) \|_2 > \tau \right)
\end{equation}
where $\mathbb{I}(\cdot)$ is the indicator function and $\tau$ is a distance threshold. The final combined mask $CM_t$ is obtained by the logical union of the semantic mask $M_t$ and the geometric mask $FM_t$, expressed as $CM_t = M_t \lor FM_t$.
This hybrid strategy ensures that pixels violating the static scene assumption—detected via semantic priors or cross-frame geometric inconsistencies—are consistently identified as dynamic regions for the subsequent rendering stage.\\

\noindent \textbf{Hybrid Reprojection.}
To synthesize the stabilized frame $S_t$, we represent the scene as a composite 3D point cloud. Based on the mask $CM_t$, we distinguish between static and dynamic regions to aggregate a global-local point set $\mathcal{P}_t$:
\begin{equation}
    \mathcal{P}_t = \mathcal{P}_t^{\text{s}} \cup \mathcal{P}_t^{\text{d}}
\end{equation}
where the static component $\mathcal{P}_t^{\text{s}} = \{ P_i(\mathbf{p}) \mid i \in \mathcal{N}(t), CM_i(\mathbf{p}) = 0 \}$ aggregates spatial information across a temporal window $\mathcal{N}(t)$ of size $n$ to fill disocclusion gaps and suppress artifacts through multi-view consistency. Conversely, the dynamic component $\mathcal{P}_t^{\text{d}} = \{ P_t(\mathbf{p}) \mid CM_t(\mathbf{p}) = 1 \}$ is restricted to the current frame $t$ to preserve the temporal integrity of non-rigid motion. The final stabilized frame $S_t$ is rendered by projecting the composite point cloud $\mathcal{P}_t$ using the smoothed camera pose $\hat{\mathbf{g}}_t$:
\begin{equation}
    S_t = \mathcal{R}(\mathcal{P}_t, \hat{\mathbf{g}}_t, K)
\end{equation}
where $\mathcal{R}(\cdot)$ denotes the point-based rendering operation implemented via PyTorch3D~\cite{ravi2020accelerating}, and $K$ represents the camera intrinsic matrix.

This hybrid stabilized rendering isolates dynamic and static points using geometry and semantics, aggregating multi-frame point clouds to fill parallax holes and reduce jitter. HSR is continuously motion-adaptive: it seamlessly transitions toward single-frame rendering for challenging motions to ensure robustness despite the risk of temporal jitter, and toward multi-frame rendering for simpler motions to maintain consistency. This mechanism guarantees semantically complete, as well as geometrically and temporally consistent rendering conditions under any extreme motion. Nevertheless, this inevitably leads to parallax occlusion holes and black borders caused by pose changes, which cannot be completely filled during the rendering stage. Furthermore, even with multi-frame alignment, depth discrepancies in single-frame rendering still introduce a certain degree of temporal jitter. These challenges are addressed in the subsequent section, which introduces a VSDM to complete and optimize the geometry-preserved frames $S_t$.
\begin{figure*}[h]
    \centering
    \includegraphics[width=0.8\textwidth]{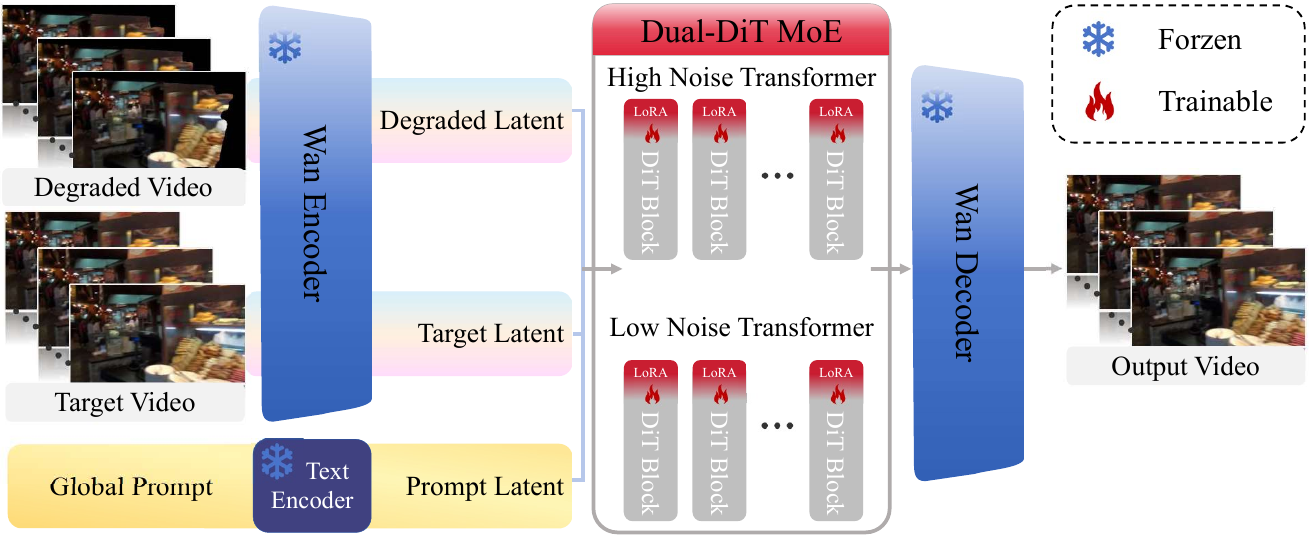}
\caption{
\textbf{Architecture of the Video Stabilization-Driven Diffusion Model.} To facilitate efficient fine-tuning, we freeze all network parameters except for the LoRA weights. Specifically, LoRA layers with a rank of 32 are integrated into every transformer block within the two DiT models.
    }  
    \label{fig:VDM}
\end{figure*}
\subsection{Full-frame Completion and Refinement}
\label{subsec:DVDM}
While the hybrid reprojection module produces geometrically stable frames $S_t$, the sequence often exhibits cropping artifacts, occlusion holes, and rendering noise. It is important to emphasize that we consider video stabilization to be an \textbf{information-overdetermined task} as shown in ~\cref{fig:neo_our3} and \cref{sec:discussion}. For parallax occlusions and black borders caused by pose transformations, the vast majority of missing information can be filled by extracting details from adjacent frames, rather than relying heavily on pure generation. Therefore, we decouple the objective of achieving full-frame reconstruction into two sub-tasks: first, fusing information from neighboring frames to fill parallax occlusions and black borders; and second, jointly optimizing texture while maintaining temporal consistency. 
To address these two tasks, we propose a \textbf{Video Stabilization-Driven Diffusion Model (VSDM)}. Formulated as a Mixture of Experts (MoE) architecture, this model comprises a high-noise Diffusion Transformer (DiT) and a low-noise DiT. During the high-noise stage, the model integrates temporal information to fill occlusion holes, while in the low-noise stage, it jointly optimizes texture and consistency. Throughout this entire process, the model effectively stabilizes temporal jitter. We intentionally suppress the model's generative capacity, while retaining its ability to synthesize reasonable content based on scene context only when small regions suffer from a complete loss of information.

The model incorporates two distinct input streams. First, a Video Conditioning Stream utilizes the rendered frames $\{S_t\}_{t=1}^N$ to provide spatial priors and motion trajectories. Instead of using standard conditional inputs, we concatenate them along the channel dimension and set the mask to all ones to denote a conditioning reference. As discussed in ~\cref{sec:discussion}, this design enables our VSDM to function as a plug-and-play module that can be applied to the output of any video stabilization method for plausible full-frame completion. Second, a Global Semantic Stream employs a fixed, universal text embedding to serve as a semantic anchor, guiding the model to maintain consistent visual quality and aesthetic style across diverse scenes. By fusing these streams, the model effectively borrows information from neighboring frames to fill disocclusion holes and rectify rendering noise, ultimately synthesizing a temporally coherent, full-frame video $\{\hat{S}_t\}_{t=1}^N$. \\

\noindent \textbf{Data Curation.} To construct a task-specific training corpus for video stabilization restoration, we generate paired data from the public MotionStab dataset~\cite{Zhang_2023_ICCV}, which comprises 220 stable-unstable video pairs across five scenarios: Regular, Crowd, Parallax, QuickRotation, and Zooming. We use the original stable videos as ground truth ($V_{target}$). To synthesize degraded videos ($V_{degrade}$), we reconstruct each stable video into frame-wise camera poses, depth maps, and RGB-D point clouds, then reproject them to their original viewpoints. To mimic stabilization artifacts, we apply depth smoothing before rendering to simulate flying points, and introduce slight camera pose jitter with sparse ray-direction perturbations to simulate HSR-induced temporal jitter under extreme motion. Additionally, we extract visibility masks by stabilizing the corresponding unstable videos, superimposing them onto the rendered frames to replicate cropping artifacts and occlusion holes from motion compensation. Ultimately, the degraded videos serve as model inputs, supervised by the original stable videos during fine-tuning. \\

\noindent \textbf{Fine-tuning.}
Our restoration model is built upon the Wan2.2-I2V-A14B framework~\cite{wan2025wanopenadvancedlargescale}, featuring a Dual-DiT Mixture-of-Experts (MoE) architecture as shown in Fig.~\ref{fig:VDM}. We employ the Wan-VAE encoder $\mathcal{E}$ to compress the degraded video $V^{degrade}$ and target video $V^{target}$ into latents $\mathbf{z}_{degrade}$ and $\mathbf{z}_{target}$, respectively. Following Flow Matching, the noisy latent $\mathbf{z}_t$ is constructed via linear interpolation:
\begin{equation}
\mathbf{z}_t = (1 - t) \epsilon + t \mathbf{z}_{target}
\end{equation}
where $t \in [0, 1]$ is the normalized timestep and $\epsilon \sim \mathcal{N}(0, \mathbf{I})$. To adapt the image-to-video setting for enhancement, we concatenate the full-sequence $\mathbf{z}_{degrade}$ with $\mathbf{z}_t$ along the channel axis to guide the restoration of artifacts and holes. During fine-tuning, the VAE is frozen, and LoRA-based adaptation is applied exclusively to the Transformer experts within the MoE structure. This preserves pre-trained generative priors while learning the mapping to clean videos. The training objective is defined by the Flow Matching loss:
\begin{equation}
\begin{split}
\mathcal{L}_{tune} &= \mathbb{E}_{\mathbf{z}_{target}, t, \epsilon, \mathbf{z}_{degrade}} \bigg[ \Big\| \mathbf{v}_\theta (\mathbf{z}_t, t, \mathbf{z}_{degrade}, c_{tex}) - \\
&\quad (\epsilon - \mathbf{z}_{target}) \Big\|_2^2 \bigg]
\end{split}
\end{equation}
where $\mathbf{v}_\theta$ is the predicted velocity and $c_{tex}$ is a global task-oriented prompt ensuring high-fidelity, artifact-free synthesis.
\\

\noindent \textbf{Implementation Details.}
\label{Implement Detail}
We fine-tune the Dual-DiT backbone of Wan2.2 via Low-Rank Adaptation (LoRA). Specifically, LoRA modules are integrated into the linear projections of both the attention mechanisms and the feed-forward networks. The model is trained on our curated dataset of noisy-clean video pairs using the AdamW optimizer with a learning rate of $\eta = 1 \times 10^{-4}$ and a weight decay of $10^{-2}$. Both the LoRA rank and alpha are set to 32. To facilitate detail restoration under high-noise regimes, we implement a shifted timestep sampling strategy with a shift factor of 5.0. The Dual-DiT MoE model is jointly trained on 4 NVIDIA H20 GPUs for 10 epochs using bf16 mixed precision, totaling approximately 560 GPU hours.
\section{Experiments}
\subsection{Experimental Setting}
\label{sec:data-setting}
\noindent \textbf{Datasets.} We evaluate our method on public dataset NUS~\cite{liu2013bundled}. The NUS dataset comprises 144 videos, categorized into six different scenes: Regular, Running, Crowd, Parallax, QuickRotation, and Running. Furthermore, we evaluate on the DeepStab~\cite{wang2018deep} dataset to confirm our model's robust generalization. \\

\noindent \textbf{Baseline.} 
We choose various video stabilization algorithms as the baselines, including RobustL1~\cite{grundmann2011auto}, Bundled~\cite{liu2013bundled}, Yu et al.~\cite{Yu_2020_CVPR}, DIFRINT~\cite{DIFRINT}, DUT~\cite{xu2022dut}, Rstab~\cite{peng20243d} and GaVS~\cite{you2025gavs}. For comparisons, we use the videos generated by official implementations with default parameters or pre-trained models. Furthermore, we compare our approach with NeoVerse~\cite{yang2026neoverse}, a state-of-the-art novel view synthesis method. We evaluate it using its highest configuration (full-frame, without LoRA) while applying the same stabilization and long-sequence fusion strategies as ours.\\

\noindent \textbf{Metrics.} We assess Cropping and Stability scores following the method in~\cite{DIFRINT}, alongside the Epipolar Sampson Error (ESE), Warping Error (WE)~\cite{lai2018learning}, and LPIPS~\cite{zhang2018unreasonable}:
\textbf{(1) Cropping Ratio:} Measures the remaining image area after removing non-content pixels.
\textbf{(2) Stability Score:} Assesses smoothness via the ratio of low-frequency to total motion energy.
\textbf{(3) Epipolar Sampson Error (ESE):} Quantifies geometric consistency by measuring point-to-epipolar-line distances across consecutive frames.
\textbf{(4) Warping Error (WE):} Evaluates temporal coherence using mean squared error between frames warped via dense optical flow.
\textbf{(5) LPIPS:} Measures perceptual similarity through deep feature comparison after image registration to ensure high-fidelity reconstruction.

\textbf{Notably}, we replace traditional distortion metrics with the ESE-WE-LPIPS triplet. Since global homography-based estimation is ill-suited for videos with significant scene depth and parallax, it fails to provide a reliable measure of geometric deformation in 3D-complex environments. Instead, this combination more reliably characterizes geometric consistency, temporal stability, and visual fidelity. Specifically, ESE measures the ability to handle parallax, WE assesses temporal stability, and LPIPS evaluates similarity to the original frames, strictly penalizing structural collapse and inconsistent content.

\subsection{Qualitative Results}
\begin{figure*}
    \centering
    \includegraphics[width=\textwidth]{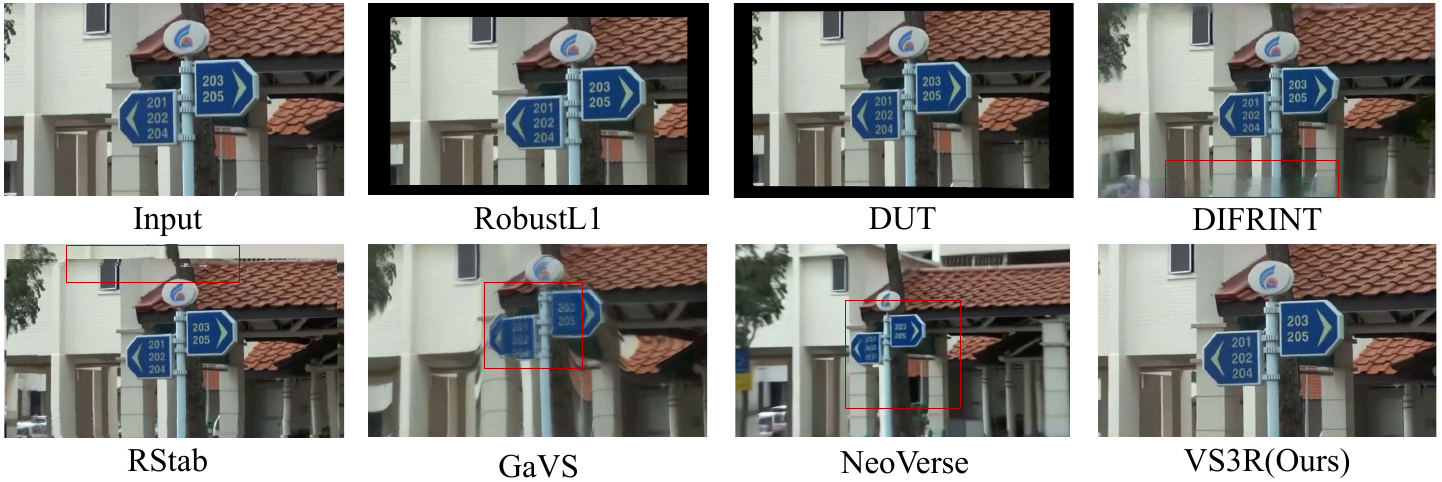}
\caption{
\textbf{Qualitative comparison of stabilized video on the test set of NUS~\cite{liu2013bundled} dataset}. Our VS3R generates stabilized videos with significantly higher content, geometric, and temporal consistency compared to existing state-of-the-art methods, including DIFRINT~\cite{DIFRINT}, Rstab~\cite{peng20243d}, GaVS~\cite{you2025gavs} and NeoVerse~\cite{yang2026neoverse}.
    }  
    \label{fig:visual_result}
\end{figure*}

Fig.~\ref{fig:visual_result} illustrates representative artifacts, distortions, and algorithmic failures observed across different methods. Due to their 2D full-frame completion paradigms, \textbf{DIFRINT} and \textbf{Rstab} suffer from significant blurring and distortions at frame boundaries under rapid camera motion, where aggressive cropping is required. While \textbf{Gavs} attempts to address these issues via 2D outpainting and 3D rendering, it is often hindered by the limitations of traditional Structure-from-Motion (SfM) preprocessing, leading to performance degradation in extreme cases such as \textit{quick rotation} and \textit{zooming}. Although \textbf{NeoVerse} achieves accurate initialization via feed-forward 4DGS, it is constrained by the rendering limitations of 3DGS under drastic focal length variations. It fails to optimize the scene to a correct depth; consequently, rendering with stabilized camera poses fails due to parallax artifacts induced by pose shifts, as shown in Fig.~\ref{fig:neo_our1}. Furthermore, as illustrated in ~\cref{fig:neo_our2,fig:neo_our3}, its over-reliance on generative priors leads to the creation of content entirely inconsistent with the original scene. In contrast, our \textcolor{blue}{VS3R} method consistently generates stable frames with significantly more complete structures and fewer severe distortions in these complex and challenging scenarios. 

\subsection{Quantitative Evaluation}
\label{subsec:quantitative}
\begin{table*}[h]
    \centering
    \caption{Quantitative comparison of video stabilization on the NUS~\cite{liu2013bundled} and DeepStab~\cite{wang2018deep} datasets. We evaluate our method against baselines using five metrics: Cropping, Stability, LPIPS, Epipolar Sampson Error(ESE), Warping Error(WE). The \textcolor{green}{best} and \textcolor{blue}{second best} results are highlighted in green and blue, respectively.}
    \resizebox{\textwidth}{!}{
    \begin{tabular}{l|c|c|ccccc|ccccc|ccccc}
        \hline
        \multirow{2}{*}{Dataset} & \multirow{2}{*}{Type} & \multirow{2}{*}{Method} & \multicolumn{15}{c}{\textbf{Average of All Categories}}\\
        \cline{4-18}
        & & & \multicolumn{3}{c}{Cropping $\uparrow$} & \multicolumn{3}{c}{Stability $\uparrow$} & \multicolumn{3}{c}{LPIPS $\downarrow$} & \multicolumn{3}{c}{ESE ($pixel^2$)$\downarrow$} & \multicolumn{3}{c}{WE($\times 1e-3$) $\downarrow$}\\
        \hline
        
        \multirow{8}{*}{NUS} & \multirow{4}{*}{Normal}
        & Bundled~\cite{liu2013bundled}  
        & \multicolumn{3}{c}{0.974} & \multicolumn{3}{c}{0.862} & \multicolumn{3}{c}{0.136} & \multicolumn{3}{c}{0.594} & \multicolumn{3}{c}{1.387} \\
        
        & & RobustL1~\cite{grundmann2011auto} 
        & \multicolumn{3}{c}{0.816} & \multicolumn{3}{c}{0.843} & \multicolumn{3}{c}{\textcolor{blue}{0.113}} & \multicolumn{3}{c}{0.361} & \multicolumn{3}{c}{\textcolor{blue}{1.064}} \\
        
        & & Yu et al.~\cite{Yu_2020_CVPR}
        & \multicolumn{3}{c}{0.877} & \multicolumn{3}{c}{0.863} & \multicolumn{3}{c}{0.142} & \multicolumn{3}{c}{0.553} & \multicolumn{3}{c}{1.279} \\
        
        & & DUT~\cite{xu2022dut} 
        & \multicolumn{3}{c}{0.877} & \multicolumn{3}{c}{\textcolor{blue}{0.894}} & \multicolumn{3}{c}{\textcolor{green}{0.107}} & \multicolumn{3}{c}{0.410} & \multicolumn{3}{c}{1.104} \\ 
        
        \cline{2-18} 
        
        & \multirow{4}{*}{Full-frame}
        & DIFRINT~\cite{DIFRINT} 
        & \multicolumn{3}{c}{0.983} & \multicolumn{3}{c}{0.848} & \multicolumn{3}{c}{0.132} & \multicolumn{3}{c}{0.956} & \multicolumn{3}{c}{1.272} \\
        
        & & Rstab~\cite{peng20243d} 
        & \multicolumn{3}{c}{0.989} & \multicolumn{3}{c}{0.866} & \multicolumn{3}{c}{0.134} & \multicolumn{3}{c}{0.369} & \multicolumn{3}{c}{1.145} \\
        
        & & GaVS~\cite{you2025gavs} 
        & \multicolumn{3}{c}{\textcolor{blue}{0.999}} & \multicolumn{3}{c}{0.841} & \multicolumn{3}{c}{0.193} & \multicolumn{3}{c}{4.136} & \multicolumn{3}{c}{1.149} \\

        & & NeoVerse~\cite{yang2026neoverse} 
        & \multicolumn{3}{c}{\textcolor{green}{1.000}} & \multicolumn{3}{c}{0.892} & \multicolumn{3}{c}{0.181} & \multicolumn{3}{c}{\textcolor{blue}{0.283}} & \multicolumn{3}{c}{1.110} \\
        
        & & VS3R (Ours) 
        & \multicolumn{3}{c}{\textcolor{green}{1.000}} & \multicolumn{3}{c}{\textcolor{green}{0.897}} & \multicolumn{3}{c}{0.176} & \multicolumn{3}{c}{\textcolor{green}{0.254}} & \multicolumn{3}{c}{\textcolor{green}{0.991}} \\
        \hline
        \hline
        \multirow{4}{*}{Deepstab} & \multirow{4}{*}{Full-frame} 
        & DIFRINT 
        & \multicolumn{3}{c}{0.982} & \multicolumn{3}{c}{0.782} & \multicolumn{3}{c}{\textcolor{blue}{0.110}} & \multicolumn{3}{c}{8.69} & \multicolumn{3}{c}{1.007} \\
        
        & & Rstab 
        & \multicolumn{3}{c}{\textcolor{green}{1.000}} & \multicolumn{3}{c}{0.834} & \multicolumn{3}{c}{\textcolor{green}{0.121}} & \multicolumn{3}{c}{0.747} & \multicolumn{3}{c}{\textcolor{blue}{0.778}} \\
        
        & & GaVS 
        & \multicolumn{3}{c}{\textcolor{green}{1.000}} & \multicolumn{3}{c}{0.812} & \multicolumn{3}{c}{0.181} & \multicolumn{3}{c}{0.928} & \multicolumn{3}{c}{0.907} \\

        & & NeoVerse
        & \multicolumn{3}{c}{\textcolor{green}{1.000}} & \multicolumn{3}{c}{\textcolor{blue}{0.861}} & \multicolumn{3}{c}{0.223} & \multicolumn{3}{c}{\textcolor{green}{0.530}} & \multicolumn{3}{c}{0.700} \\
        
        & & VS3R (Ours) 
        & \multicolumn{3}{c}{\textcolor{green}{1.000}} & \multicolumn{3}{c}{\textcolor{green}{0.863}} & \multicolumn{3}{c}{0.210} & \multicolumn{3}{c}{\textcolor{blue}{0.589}} & \multicolumn{3}{c}{\textcolor{green}{0.571}} \\
        \hline
    \end{tabular}
    \label{tab:all_results1}
    }
\end{table*}
Tab.~\ref{tab:all_results1} presents the average results for six categories on the NUS dataset. As shown in \cref{tab:all_results1}, \textcolor{blue}{VS3R} achieves significant performance improvements over existing video stabilization methods.

Regarding stability, our method's tolerance to cropping allows for larger smoothing bandwidths, which leads to superior stability. In terms of consistency, our approach outperforms other methods in both geometric and temporal coherence. 3D reconstruction-based methods often lag behind 2D methods in the LPIPS metric. However, this indirectly demonstrates the effectiveness of the 3D stabilization paradigm in handling parallax. First, transformations from pose optimization alter occlusion relationships, generating valid novel content absent from the original images. The LPIPS metric misinterprets this valid new content as image distortion. Second, the 3D re-rendering process inevitably introduces minor pixel-level residuals. Nevertheless, a reference-based similarity metric remains necessary. It provides a critical penalty for structural collapse or the generation of content entirely inconsistent with the original image. As part of a joint evaluation framework, it must be considered alongside the other two metrics to properly assess the quality of the stabilized images. Furthermore, our user study confirms these shifts do not affect visual quality, representing a necessary trade-off for superior 3D consistency.

\subsection{User Study}
\label{sec:user_study}
\begin{figure}
    \centering
    \includegraphics[width=\columnwidth]{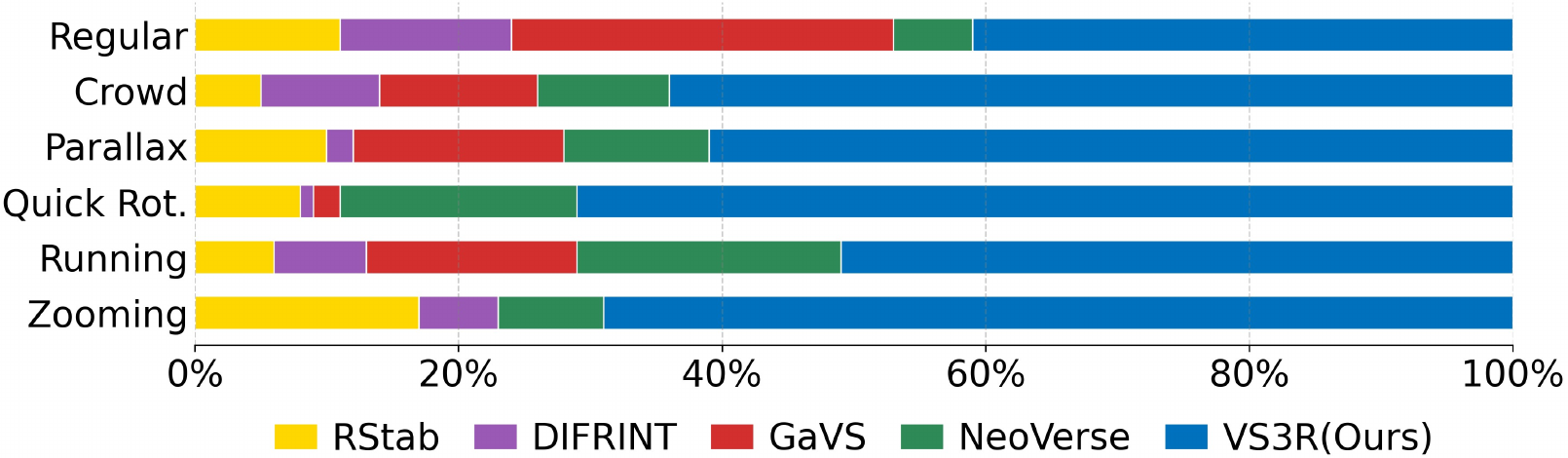}
\caption{
User preference test results on 6 video categories.
    }  
    \label{fig:user_study}
\end{figure}

We conducted a user study to compare our method with several state-of-the-art full-frame approaches, namely DIFRINT~\cite{DIFRINT}, Rstab~\cite{peng20243d}, NeoVerse~\cite{yang2026neoverse} and Gavs~\cite{you2025gavs}. Specifically, we prepared a test set consisting of 24 video sequences (4 videos per category). We recruited 25 participants for a blind user study. For each test case, participants were presented with the original unstable video as a reference and asked to select the most visually pleasing result among the five randomized competing methods. The statistical results of this user study are illustrated in Fig.~\ref{fig:user_study}, demonstrating that our method is consistently preferred by users.

\subsection{Ablation Study}
\label{subsec:ablation_study}
\begin{figure*}
    \centering
    \includegraphics[width=0.9\textwidth]{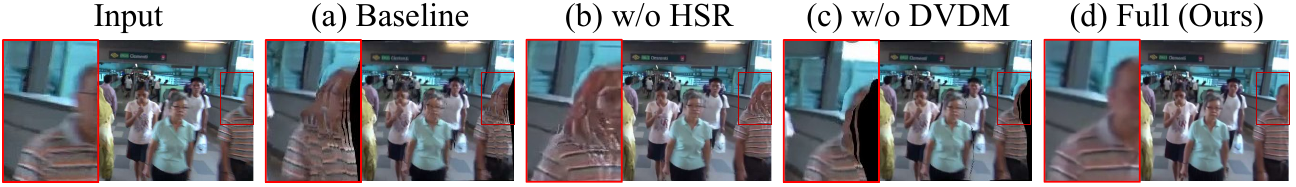}
\caption{
\textbf{Qualitative ablation of HSR and VSDM.} (b) w/o HSR: Artifacts and structural instability arise from relying solely on predicted masks. (c) w/o VSDM: Results exhibit disocclusion holes and texture loss without diffusion-based completion. (d) Full (Ours): HSR ensures geometric stability in dynamic regions, while VSDM restores disoccluded areas with temporal coherence and high fidelity.}
    \label{fig:ablation_study}
\end{figure*}
We evaluate the contributions of the Hybrid Stabilized Rendering (HSR) and Video Stabilization-Driven Diffusion Model (VSDM) modules. Experiments are conducted on the dynamic Crowd scene from the NUS dataset~\cite{liu2013bundled}, selected for its complex motions and dynamic environments.\\

\noindent \textbf{Effectiveness of HSR.}
We validate HSR by comparing it against a baseline(Fig.~\ref{fig:ablation_study}\textcolor{red}{a}) that renders solely on raw dynamic masks $M_t$ from VGGT4D~\cite{hu2025vggt4d}. As illustrated in Fig.~\ref{fig:ablation_study}\textcolor{red}{a} and Tab.~\ref{tab:ablation_metrics}, inaccuracies in network-predicted masks lead to significant rendering artifacts. By integrating semantic-driven mask $M_t$ with geometric mask $FM_t$, HSR effectively mitigates these errors, maintaining structural integrity and suppressing artifacts in dynamic regions.\\

\noindent \textbf{Effectiveness of VSDM.}
The role of VSDM is analyzed by excluding the diffusion-based refinement process. Without VSDM, the rendered frames suffer from disocclusion holes and missing textures as shown in Fig.~\ref{fig:ablation_study}\textcolor{red}{b} and Tab.~\ref{tab:ablation_metrics}, as geometric projection alone cannot synthesize unseen content. In contrast, VSDM leverages its MoE architecture to fill these gaps while enforcing temporal consistency. Results demonstrate that VSDM significantly enhances perceptual quality, producing photo-realistic stabilized videos.
\begin{table*}[htb]
    \centering
    \caption{\textbf{Quantitative ablation study of HSR and VSDM modules}. We conduct comparative experiments of the effect of each module.The \textcolor{green}{best} and \textcolor{blue}{second best} results are highlighted in green and blue, respectively.}
    \resizebox{0.8\textwidth}{!}{
    \begin{tabular}{cccccc}
        \toprule
        Method & Cropping $\uparrow$ & Stability $\uparrow$ & LPIPS $\downarrow$ & ESE ($pixel^2$)$\downarrow$ & WE($\times 1e-3$) $\downarrow$\\
        \midrule
        Baseline & 0.994 & 0.874 & 0.231 & 1.361 & 1.383\\
        w/o HSR & \textcolor{green}{1.000} & \textcolor{blue}{0.892} & \textcolor{blue}{0.197} & \textcolor{blue}{0.586} & 1.135 \\
        w/o VSDM & 0.987 & 0.887 & 0.214 & 0.675 & \textcolor{blue}{1.128} \\
        Full(Ours) & \textcolor{green}{1.000} & \textcolor{green}{0.897} & \textcolor{green}{0.176} & \textcolor{green}{0.254} & \textcolor{green}{0.991} \\
        \bottomrule
    \end{tabular}%
    \label{tab:ablation_metrics}
    }
\end{table*}
\section{Discussion}
\label{sec:discussion}
\noindent $\bullet$ \textbf{\textit{Why deep 3DR?}} 
Compared to traditional Structure-from-Motion (SfM), joint multi-frame reconstruction based on feed-forward neural networks provides more robust camera pose and depth estimation. Additionally, it avoids the heavy offline preprocessing typically required by monocular methods to resolve unnormalized translation scales using target poses and point clouds. Furthermore, monocular depth estimation often introduces severe flying pixels at depth discontinuities, hindering the proper handling of occlusion relationships.

\begin{figure*}[htbp]
    \centering
    
    \begin{subfigure}[b]{\textwidth}
        \centering
        \includegraphics[width=\textwidth]{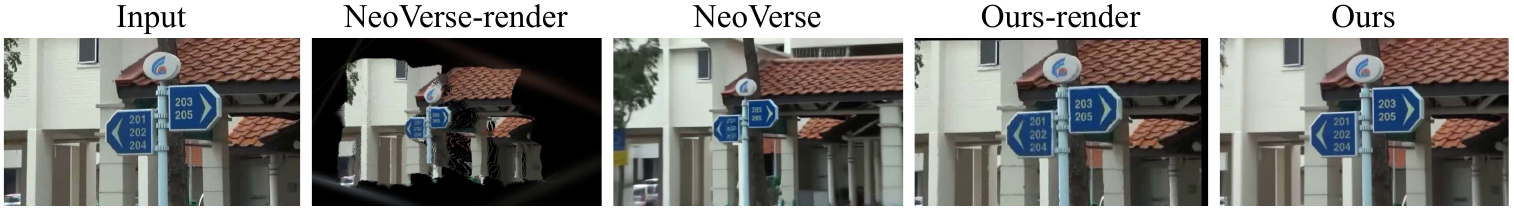}
        \caption{Drastic focal length variations.}
        \label{fig:neo_our1}
    \end{subfigure}
    
    \begin{subfigure}[b]{\textwidth}
        \centering
        \includegraphics[width=\textwidth]{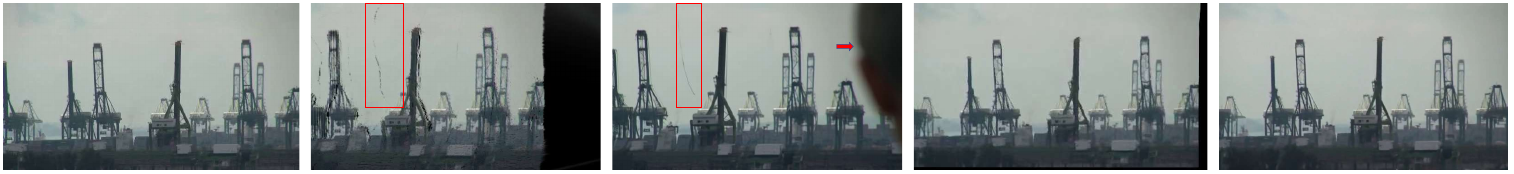}
        \caption{Artifacts and irrelevant generation}
        \label{fig:neo_our2}
    \end{subfigure}

    \begin{subfigure}[b]{\textwidth}
        \centering
        \includegraphics[width=\textwidth]{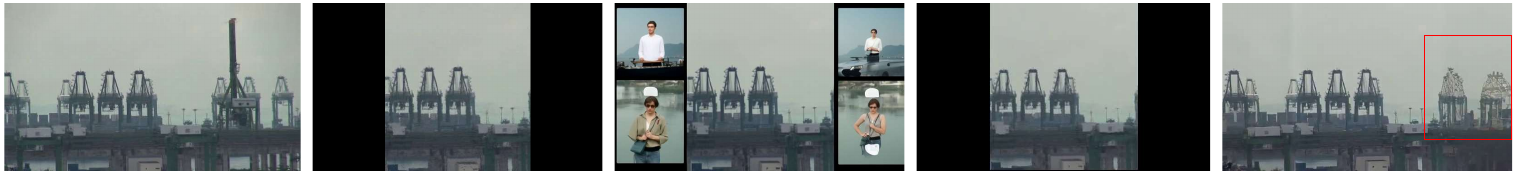}
        \caption{Extreme information missing experiments}
        \label{fig:neo_our3}
    \end{subfigure}
    
    \caption{
        A comparison with NeoVerse. \textbf{\textcolor{blue}{We provide video demonstrations in the supplementary materials}}. (a) Drastic focal length variations: Even with good initialization, the Gaussian reconstruction module of NeoVerse fails to handle degraded scenes, such as drastic focal length changes, due to its inherent scale-blindness. (b) Artifacts and irrelevant generation: It suffers from severe aliasing and scale-blindness artifacts, alongside scene-irrelevant content produced by the generative model. These artifacts are widespread across various scenes and are even more severe than depicted in the figures, specifically manifesting as glitches and strip-like cracks. The irrelevant generated content mostly appears within occlusion holes and black borders caused by camera pose shifts under large parallax. (c) Extreme information missing experiments.
    }
    \label{fig:compare_with_neoverse}
\end{figure*}

\noindent $\bullet$ \textbf{\textit{Why is video stabilization an information-overdetermined task?}} 
Although 3D-based methods like Rstab can achieve full-frame stabilization without generative models, they often suffer from severe misalignments in occlusion and border regions. To demonstrate this information-overdetermined nature, we conduct an extreme experiment by artificially masking 25\% of the visual conditions on both the left and right sides of the diffusion model input. For the NeoVerse baseline, these masked regions are padded with actual black border values to ensure semantic consistency among the input control conditions. As shown in \cref{fig:neo_our3} and the supplementary video, even with 50\% of the spatial information discarded, scene consistency is successfully maintained by fusing adjacent frames, requiring minimal purely generated content. This confirms that even severe cropping from extreme camera motion can be compensated by the abundant information inherently present in neighboring frames. Thus, the core challenge in generative full-frame stabilization lies in robustly extracting and integrating this adjacent-frame information while strictly preserving geometric and temporal consistency.

\noindent $\bullet$ \textbf{\textit{Applications?}} 
Leveraging a single-RGB concatenation input derived from our deep insights into video stabilization, VSDM can serve as a standalone module to perform full-frame restoration and temporal stabilization on the outputs of various non-full-frame methods as shown in Appendix Fig. \textcolor{red}{1}. Furthermore, our method is readily applicable to video outpainting tasks as shown in Appendix Fig. \textcolor{red}{2}. Finally, by decoupling motion from geometry, it preserves visual consistency and flexibly renders outputs for various camera models, as shown in Appendix Fig. \textcolor{red}{3}.

\noindent $\bullet$ \textbf{\textit{Limitation?}}
Despite its effectiveness, \textcolor{blue}{VS3R} has limitations. 
\textbf{\textit{(1)}}Reliance on Deep 3DR: Our framework relies on robust poses and accurate depth. However, intense depth fluctuations can introduce temporal jitter, a characteristic challenge for 3D stabilization in highly dynamic scenes. 
\textbf{\textit{(2)}}Fidelity Constraints: Pre-trained diffusion models may occasionally degrade fine textures. Though perceptual impact is minimal, future foundation models will improve output precision. 
\textbf{\textit{(3)}}Computational Overhead: Achieving state-of-the-art full-frame consistency demands substantial resources. While inference time is on par with 3D stabilizers, our framework requires higher VRAM (Appendix Sec. B). This memory trade-off is justified by our strictly superior stabilization and visual quality. Future optimizations will explore lightweight diffusion models and efficient sampling to reduce this overhead.

\section{Conclusion}
In this paper, we introduced \textcolor{blue}{VS3R}, a robust video stabilization framework that effectively addresses the fundamental trade-off between geometric robustness and full-frame consistency. By synergizing deep 3D reconstruction with generative video diffusion, our approach transcends the limitations of traditional 2D warping and fragile SfM-based optimization pipelines. Through a feed-forward reconstruction network and a \textbf{HSR} module, we ensure geometric fidelity and dynamic consistency even in challenging, unconstrained environments. Furthermore, our \textbf{VSDM} enables high-fidelity full-frame restoration by rectifying projection artifacts and disocclusion holes without cropping. Extensive qualitative and quantitative evaluations, supported by a user study, demonstrate that VS3R significantly outperforms state-of-the-art methods.

\bibliographystyle{IEEEtran}
\bibliography{main}

\end{document}